\documentclass[conference]{IEEEtran}
\pdfoutput=1

\usepackage[
backend=biber,
sorting=ynt
]{biblatex}
\usepackage{graphicx}
\usepackage{float}
\usepackage[ruled,vlined,linesnumbered]{algorithm2e}
\usepackage{amsmath}
\usepackage{algpseudocode}
\usepackage{subfigure}
\usepackage{amsthm}
\usepackage{fancybox}
\usepackage{booktabs}
\usepackage[parfill]{parskip}
\usepackage{array,booktabs,multirow}
\usepackage{textcomp}
\usepackage{url}

\usepackage[utf8]{inputenc} 
\usepackage[T1]{fontenc}    
\usepackage{hyperref}       
\usepackage{url}            
\usepackage{booktabs}       
\usepackage{amsfonts}       
\usepackage{nicefrac}       
\usepackage{microtype}      
\usepackage{xcolor}         
\usepackage{algpseudocode}
\usepackage{graphicx}
\usepackage{float}
\hypersetup{
   breaklinks=true,   
   colorlinks=false,   
}
\title{Efficient Graph-Friendly COCO Metric Computation for Train-Time Model Evaluation}

\author{%
  Luke Wood\thanks{\url{https://lukewood.xyz}} \\
  Google \\
  \texttt{lukewood@google.com} \\
  
  \and

  Fran\c{c}ois Chollet \\
  Google \\
  \texttt{fchollet@google.com} \\
  
}

\begin{document}
\maketitle

\begin{abstract}
Evaluating the COCO mean average precision (MaP) and COCO recall metrics
as part of the static computation graph of modern deep learning frameworks poses a unique set of challenges.
These challenges include the need for maintaining a dynamic-sized state to compute mean average precision,
reliance on global dataset-level statistics to compute the metrics, and managing differing numbers of bounding boxes between images in a batch. 
As a consequence, it is common practice for researchers and practitioners to evaluate COCO metrics as a post training evaluation step. 
With a graph-friendly algorithm to compute COCO Mean Average Precision and recall, these metrics could be evaluated at training time,
improving visibility into the evolution of the metrics through training curve plots,
and decreasing iteration time when prototyping new model versions.

Our contributions include an accurate approximation algorithm for Mean Average Precision, an open source implementation of both COCO mean average precision and COCO recall, extensive numerical benchmarks to verify the accuracy of our implementations, and an open-source training loop that include train-time evaluation of mean average precision and recall.
\end{abstract}

\section{Introduction}
COCO metrics, originally proposed alongside the COCO dataset, have become the
evaluation method of choice for object detection, segmentation map, and keypoint
detection models \cite{coco}.
The process of computing COCO metrics is complex and does not cleanly fit
into the computation model used in popular static graph-based deep learning frameworks such as
TensorFlow \cite{tensorflow} and JAX \cite{jax}.  For these reasons,
researchers and practitioners alike most commonly perform COCO metric evaluation as a
post-training evaluation step using the \texttt{pycocotools} package \cite{coco} \footnote{
    pyococotools is open source on GitHub at \raggedright{
        \hyperlink{https://github.com/cocodataset/cocoapi/tree/master/PythonAPI/pycocotools}{https://github.com/cocodataset/cocoapi/tree/master/PythonAPI/pycocotools}
    }
}
As a consequence, model comparison is done using a table of single scalar metrics rather than using learning curve plots.
This decreases visibility into the evolution of these metrics,
greatly increases iteration time when prototyping,
and forces comparisons across models to be based on a single number (generally without error bars),
which decreases the reliability of these comparisons.

This work contributes an approach to compute recall and mean average precision as
defined in the Microsoft COCO challenge within the constraints of the computation graphs of
TensorFlow.
Our contributions include an algorithm
that computes a close approximation to mean average precision,
fine-grained numerical accuracy benchmarks against existing COCO metric evaluation tools,
an open source Keras-native implementation of both mean average precision and recall, and an
open-source object detection pipeline leveraging these metrics.
Additionally, our algorithm and implementation support distributed metric evaluation.

These innovations enable researchers to produce learning curves for MaP and recall metrics
throughout training, which in turn enables rapid iteration, deeper
understanding of the training process, and robust result reporting.

\section{Related Work}

COCO metrics were first proposed in the Microsoft COCO challenge by Lin et al. \cite{coco}.
The metrics were used as the evaluation criteria for the challenge, and have since been the standard evaluation criteria for object detection models.  COCO metrics have been used for model evaluation in numerous works \cite{https://doi.org/10.48550/arxiv.1506.02640}\cite{rcnn}\cite{https://doi.org/10.48550/arxiv.2203.03605}\cite{https://doi.org/10.48550/arxiv.2111.09883}\cite{yolov4}\cite{yolox}\cite{retinanet}.


While COCO metrics remain the dominant object detection metrics, other object detection evaluation metrics do exist.
These include average IoU\cite{zhou2019iou}, F-Score, and traditional recall and precision.

\section{Metric Overview}
\subsection{Terminology}
\textbf{Intersection over Union (IoU).} Intersection over Union, or IoU,
is an area-based metric used in Computer Vision to determine if two objects can be considered to be the
same.  As the name suggests, the metric is
computed by dividing the area of the intersection of two objects by the area of their
union.  The range of IoU is 0 to 1, with two identical objects having an IoU of 1 and
objects with no intersection possessing an IoU of 0.

\textbf{Ground Truth.}  Predicted boxes consist of six scalars.  The first four
represent the boxes' $left$, $top$, $right$ and  $bottom$ coordinates, the fifth the
class of the predicted box, and the sixth the confidence score of the predicted box.

\textbf{Predicted Boxes.}  Predicted boxes contain the same information as ground-truth
boxes, along with an additional sixth value representing the confidence score of the
predicted box.  Predicted boxes are typically produced by a bounding box detection model.

\textbf{True Positive and False Positive.} To determine if a predicted box is a true positive or false positive, first we define some IoU threshold $\theta$.

Next, we consider the predicted and true boxes present in a single image.
We compute the IoU scores for the predicted boxes and the true boxes, iterating over the predicted boxes.
If the predicted box has an IoU larger than $\theta$ for one or more ground truth box with the same class ID, the ground truth box with
the highest IoU to the predicted box is marked as matched.
The matched predicted box is then considered a true positive, and the process continues until all predicted or
ground truth boxes are exhausted.  Any remaining predicted boxes are considered false
positives.

\subsection{Recall}
An average of many recall scores is often used to represent a single value of recall in object detection benchmarks.  These various recall scores are computed given a set of IoU thresholds, $\Theta$.  A common choice for $\Theta$,
originally used in the Microsoft COCO challenge, is the range $[0.5, 0.55, ..., 0.9, 0.95]$.

To compute each individual recall value, we compute true positives and false positives in a given image with $IoU=\theta$ for all $\theta \in \Theta$.

To compute recall, the number of true positives computed for the given IoU threshold is divided by the number of ground truth boxes.  This process is repeated for all $\theta \in \Theta$, and the average of all of these is considered the final COCO recall score.

\subsection{Mean Average Precision}
Precision is defined as the number of true positives divided by the sum of true and false positives.
While precision can be computed using the standard
precision formula, this is unfortunately not particularly useful in the object detection
domain, as models may make many predictions, some of which have low confidence scores.

MaP is accounts for this fact and serves as a strong general purpose object detection
model evaluation metric.
MaP is parameterized by a set of recall thresholds $R$, and a set of IoU thresholds
$\Theta$.
While many metrics can be computed on a per-image basis, MaP can only be
computed on a global basis.

First, we construct the product of the sets $R$ and $\Theta$. We consider each subset of these
parameters, $\gamma_j \in R$ and $\theta_i \in \Theta$.

Next, the aforementioned process of determining whether or not each bounding box is a
true positive or false positive is followed with the selected value $\theta_i$.
The global set of predicted bounding boxes from the entire evaluation dataset is
sorted by confidence.  The set of all ordered subsets of boxes is
constructed, starting from the empty set, moving to the set of only the highest
confidence box, then adding an additional box by confidence level in descending order until the set of all predicted boxes is considered.  For each
set the recall and precision scores $r_{zi}, p_{zi}$ are both computed.

Following this, all pairs $r_{zi}, p_{zi}$ are considered.  A value $m_{ji}=p_{ki}$ such that $k = min(z)$ such that $r_{zi}>=\gamma_j$.
If no value $r_{zi}>=\gamma_j$, $m_{ji}=0$.  After computing all $m_{ji}$, a curve is constructed such that
$f_i(j)=m_{ji}$.  A final value $\tau_i = \int_{0}^{1} f_i(j) \,dj $
is computed.

The above process is repeated for all $\theta_i$.  Finally, the result of MaP is defined $MaP = avg_i(\tau_i)$.

This process is complex, involves many edge cases, and relies on constructs
incompatible with the static graph-based computation model of the popular deep learning frameworks
TensorFlow.
In practice, implementations must meticulously consider
efficiency and order of operations to achieve suitable runtime performance.

\section{Graph Implementation}
\label{sec:difficulties-faced}

Implementing COCO metrics in a graph deep learning framework poses numerous challenges.
This section describes each of the challenges faced in our implementation, as well as
the techniques and algorithms we implemented to solve them.

\subsection{Variable Size State.}
The primary issue with computing MaP in the context of a computation graph is that MaP
requires a variable-size state representation.
This is because in order to compute MaP bounding box, predictions must be sorted on a
global-batch basis according to the detection confidence score.
Whether or not a detection was a true positive or false
positive must be stored internally in the metric, alongside the corresponding confidence
score.  The number of predictions for any given
dataset is unknown until the full dataset is evaluated, therefore the internal state
size of MaP is variable.  Modern deep learning frameworks lack strong support for
variable size state in their computation graphs.

\textit{MaP Approximation Algorithm.)}
In order to solve this issue we propose an algorithm to
closely approximate MaP.  Instead of storing a list of detection instances, whether or not they
were true positives, and their corresponding confidence score we initialize twos vector,
$\beta_p$ and $\beta_n$, both of shape $(buckets,)$.
These vectors are initialized to contain all zeros.

Evaluation is performed on a mini-batch basis.  During each mini-batch, we first
determine if the detection instances are true positives or false positives.
Following this, an index is computed for each bounding box using
$i=floor(c*(buckets-\delta))$, where $c=confidence$ and $\delta$ is an infinitesimal.
This yields a number in the range $[0, buckets)$.  If the bounding box is a true positive,
$\beta_p[i] = \beta_p[i]+1$, otherwise $\beta_n[i] = \beta_n[i]+1$.

Additionally, the number of ground truth boxes is also tracked in a Tensor, $\Gamma$ in order to compute the
recall scores as required during the AuC interpolation step.

At the end of evaluation of all mini-batches, the final result is computed using the
following process.

First, a cumulative sum operation is performed over $\beta_p$ and $\beta_n$.  $\beta_{ps}$=$cumsum(\beta_p)$, $\beta_{ns}$=$cumsum(\beta_n)$. $\beta_{ps}[i]$ can be interpreted as the number of true positives the classifier has scored when only considering bounding boxes with confidence score $i/buckets$.

Following this, recall scores for every bucket are computed: $R=\beta_{ps}/\Gamma$.
Precision scores for every bucket are also computed: $P=\beta{ps}/(\beta_{fs} + \beta_{ps})$.  These recall and precision values are then used in the precision recall pair matching process described in the original.
The area under curve computation process is unchanged from the original, as is the averaging and repetition of the process over each IoU threshold.

It should be noted that an additional axis may be added to $\beta_p$, $\beta_n$, and $\Gamma$ to support multi-class MaP computation.
process.

While our algorithm
Our algorithm exposes a parameter, $buckets$, to allow end users to perform a trade off between
accuracy and runtime performance.  We recommend a default setting of $buckets=10000$.  In our
benchmarks $buckets=10000$ achieves strong accuracy, while maintaining tolerable runtime performance.

\subsection{Ragged Bounding Box Tensors.} The number of bounding boxes in an image may
differ between training samples.  This makes the bounding box Tensor ragged.  While an
easily solvable problem, this is an important design consideration when implementing
COCO metrics in a computation graph.

\textit{-1 Masking.}
In order to solve this, our implementation simply ignores samples with a class ID of $-1$.  This allows us to keep Tensors dense, while still supporting a variable number of bounding boxes in each image.  One major caveat of this approach is the additional memory use required to store the $-1$ values used to represent the masked out boxes.  While this is far from ideal, all major deep learning frameworks are rapidly improving support for RaggedTensors.  As such, our implementation will be updated to support them in the future.

\section{Numerical Accuracy Benchmarking}
In order for our metrics to be useful to researchers and practitioners it is important that the metrics be accurate to their predecessor: \texttt{pycocotools}.

It should be noted that deviation from \texttt{pycocotools} should not be thought of as a deviation from the numerical value of the underlying metric itself.
\texttt{pycocotools} lacks precise testing and was primarily developed by one individual contributor. But while \texttt{pycocotools} results should not be considered the ground-truth value for MaP or recall, it is still the industry standard for COCO metric evaluation.  As such, we consider it prudent to benchmark the error margins between our implementation and the \texttt{pycocotools} implementation.

\subsection{Mirroring Standard COCO Metric Evaluation}

The standard \texttt{pycocotools} COCO evaluation suite contains several additional parameterizations
of the standard mean average precision and recall metrics.  These additional parameterizations
include area range filtering, where only bounding boxes inside of a specified area range are considered, and a max-detection value, a limit on the number of detections a model may make per image.

In order to evaluate the numerical accuracy of our implementation we have implemented
support for these parameters.

While implementing area range filtering and max detections on top of the algorithm we have proposed in \ref{sec:difficulties-faced} is trivial, there is no one-to-one perfectly matching method.  In the mini-batch update step all examples are simply
filtered based on their computed area before computing false positives, true positives, and ground truths.

\subsection{Experimental Setup}
Our benchmarking process covers each COCO metric configuration used in the original Microsoft COCO challenge.
The accuracy of each configuration is evaluated for varying numbers of images and bounding boxes.
Each test set is generated based on the ground truth annotations provided in the COCO dataset evaluation.
Our process for producing a test dataset is as follows:
First, we sample $N$ random images from the COCO dataset.  We select all of the bounding boxes associated with that dataset and treat these as our ground truths.  Next, we modify the ground truth bounding boxes by shifting the $x$ coordinate up to 10\% of the width of the box, and shifting the $y$ coordinate by up to 10\% of the height of the box.  Following this, we randomly shift the width and height of the predicted boxes by up to 10\%.  These mutated boxes are used as predictions $y_{pred}$.

More formally:

\begin{math}
left = left + random(-0.2, 0.2)*width \\
right = right + random(-0.2, 0.2)*height \\
width = width * random(0.8, 1.2) \\
height = height * random(0.8, 1.2)
\end{math}

Test datasets are generated 10 times each, and each metric configuration is evaluated independently.
Every metric configuration from the COCO standard set of metrics is evaluated, with both the \texttt{pycocotools} implementation and the corresponding \texttt{keras\_cv} implementation.  The differences are compared, and the error margin between each is disclosed in section \ref{sec:experimental-results}.

\section{Experimental Results}
\label{sec:experimental-results}

\begin{figure*}
\begin{minipage}{1.0\textwidth}
\includegraphics[width=\textwidth]{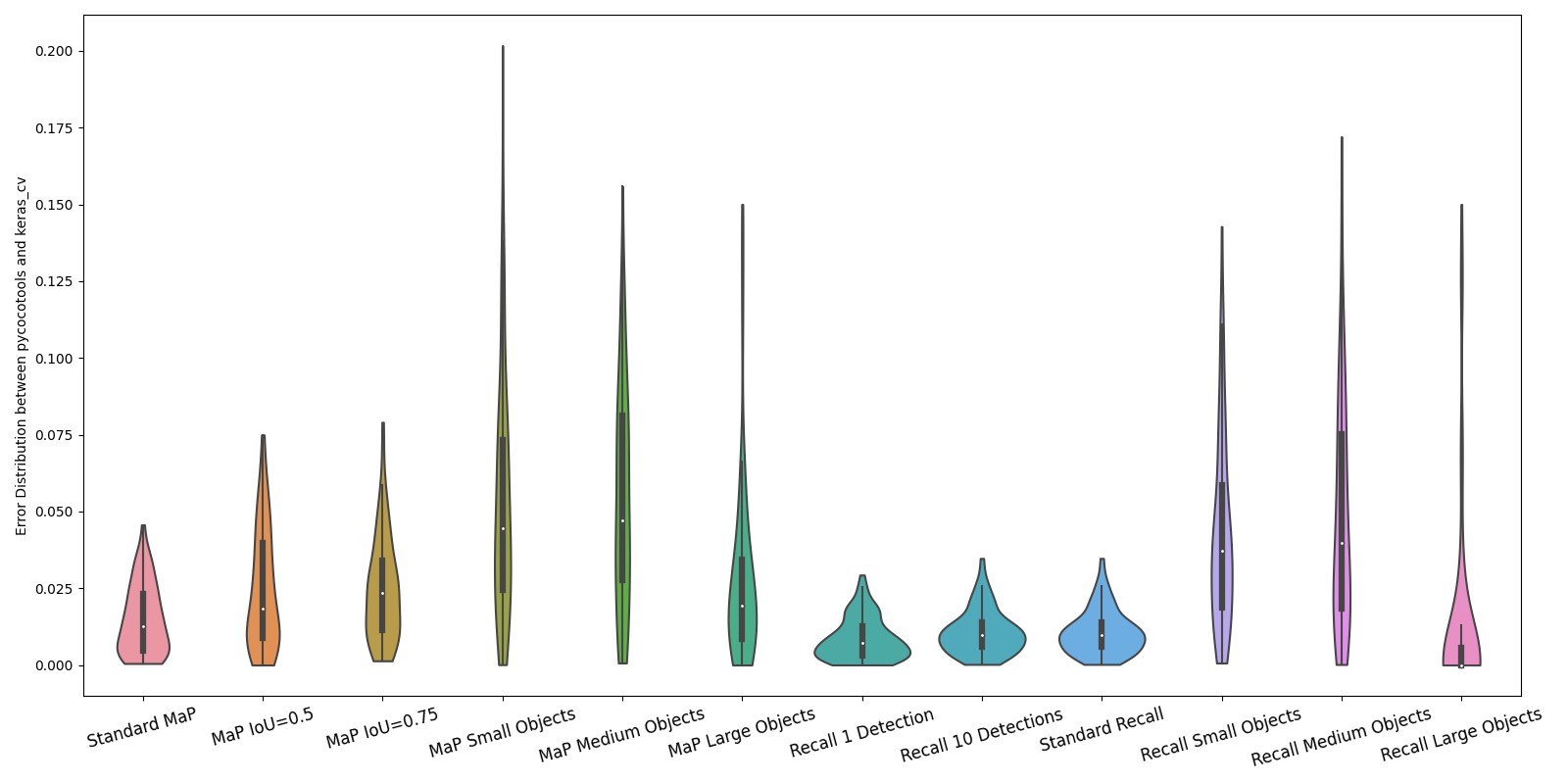}
\end{minipage}
\label{fig:violin_plot}
\caption{Error distribution for each metric configuration}
\end{figure*}

\begin{figure*}
\begin{minipage}{1.0\textwidth}
    \begin{figure}[H]
        \begin{minipage}{0.48\textwidth}
            \includegraphics[width=\textwidth]{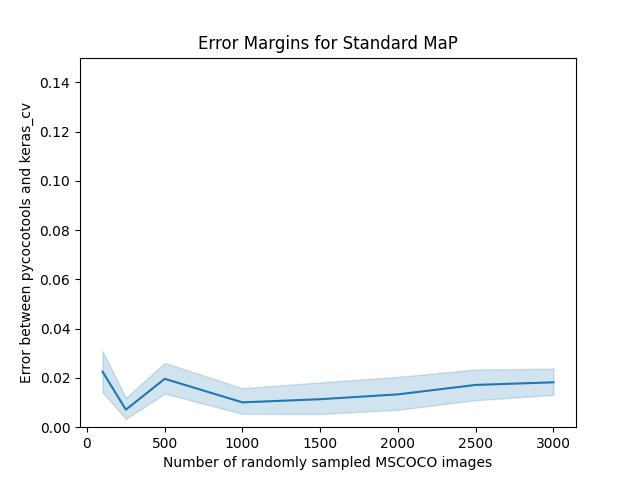}
        \end{minipage}
        \begin{minipage}{0.48\textwidth}
            \includegraphics[width=\textwidth]{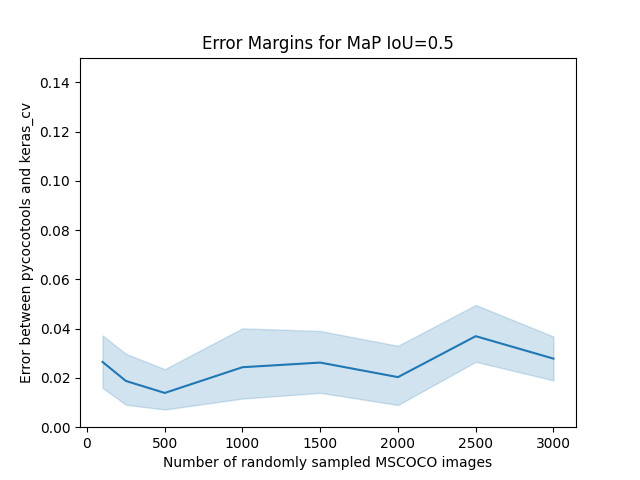}
        \end{minipage}
    \end{figure}

    \begin{figure}[H]
        \begin{minipage}{0.48\textwidth}
            \includegraphics[width=\textwidth]{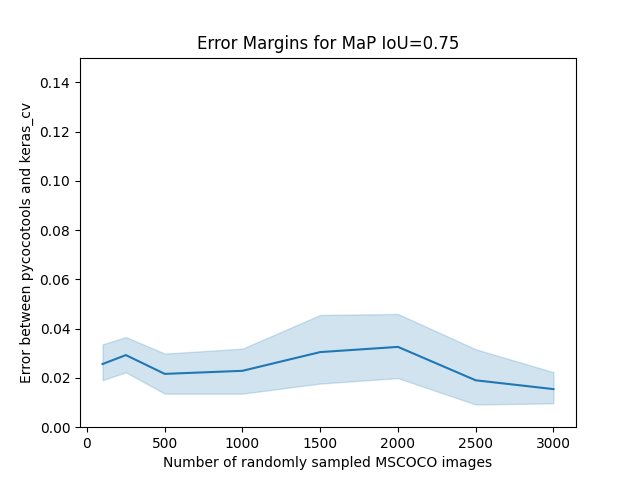}
        \end{minipage}
        \begin{minipage}{0.48\textwidth}
            \includegraphics[width=\textwidth]{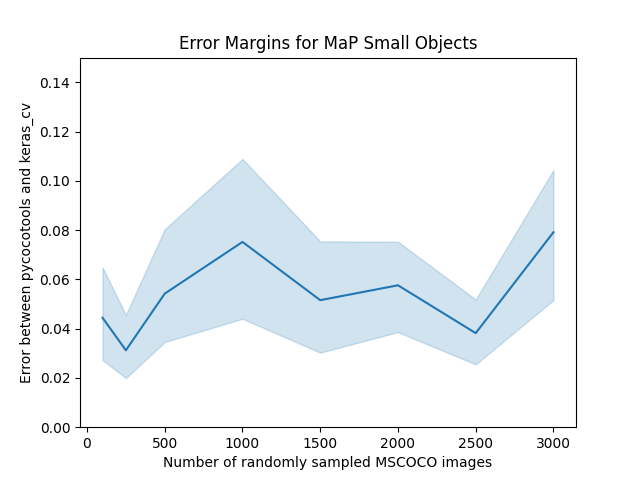}
        \end{minipage}
    \end{figure}

    \begin{figure}[H]
        \begin{minipage}{0.48\textwidth}
            \includegraphics[width=\textwidth]{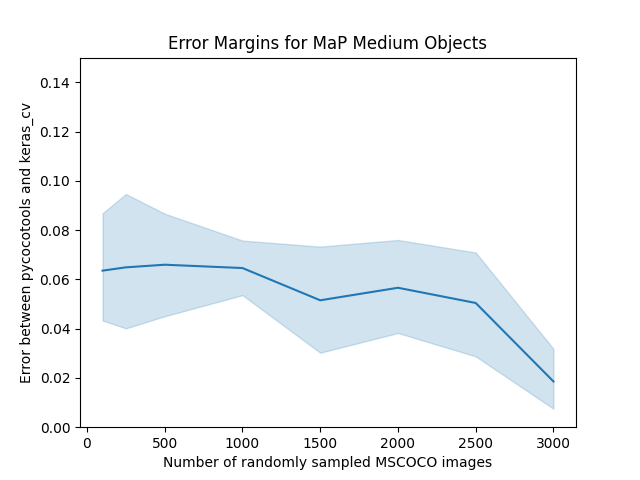}
        \end{minipage}
        \begin{minipage}{0.48\textwidth}
            \includegraphics[width=\textwidth]{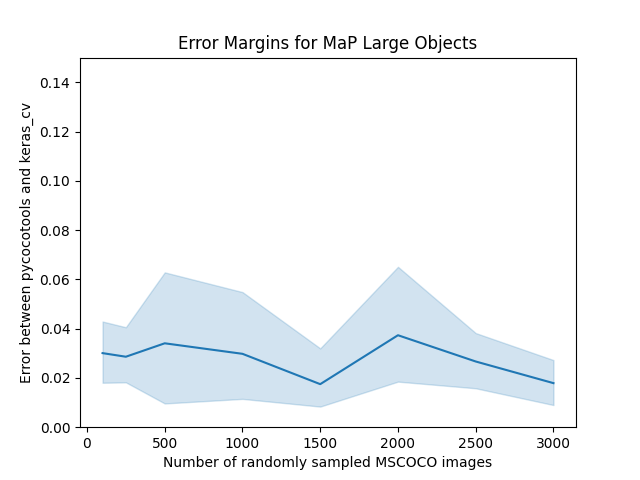}
        \end{minipage}
    \end{figure}
\end{minipage}

\caption{Error margins for each COCO mean average precision challenge configuration}
\label{fig:error-margins-map}
\end{figure*}

\begin{figure*}
\begin{minipage}{1.0\textwidth}
    \begin{figure}[H]
        \begin{minipage}{0.48\textwidth}
            \includegraphics[width=\textwidth]{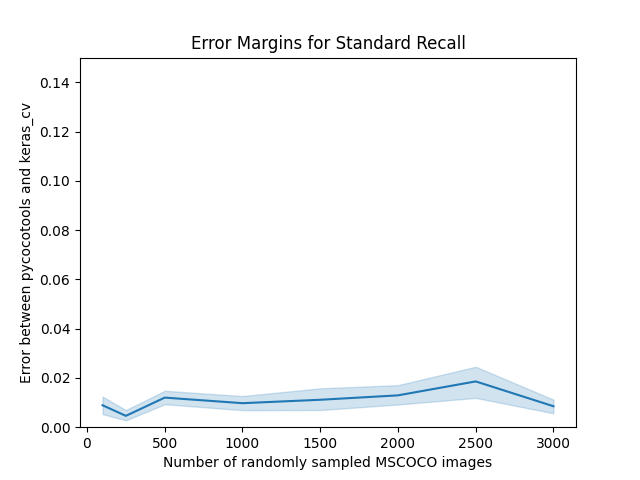}
        \end{minipage}
        \begin{minipage}{0.48\textwidth}
            \includegraphics[width=\textwidth]{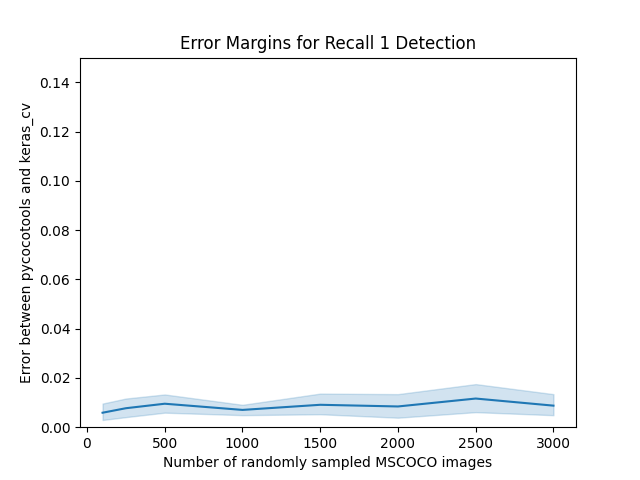}
        \end{minipage}
    \end{figure}

    \begin{figure}[H]
        \begin{minipage}{0.48\textwidth}
            \includegraphics[width=\textwidth]{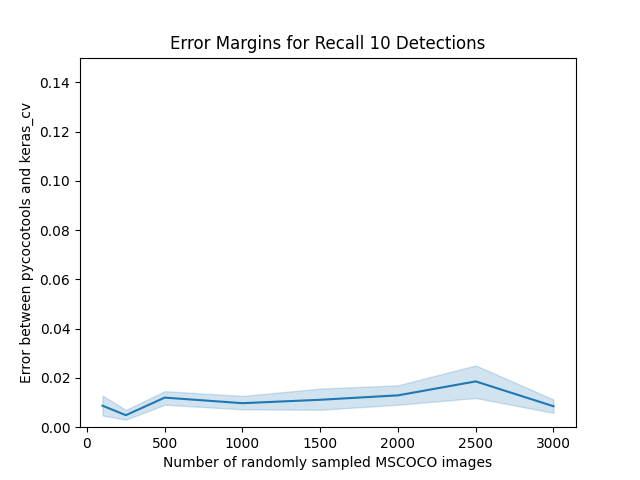}
        \end{minipage}
        \begin{minipage}{0.48\textwidth}
            \includegraphics[width=\textwidth]{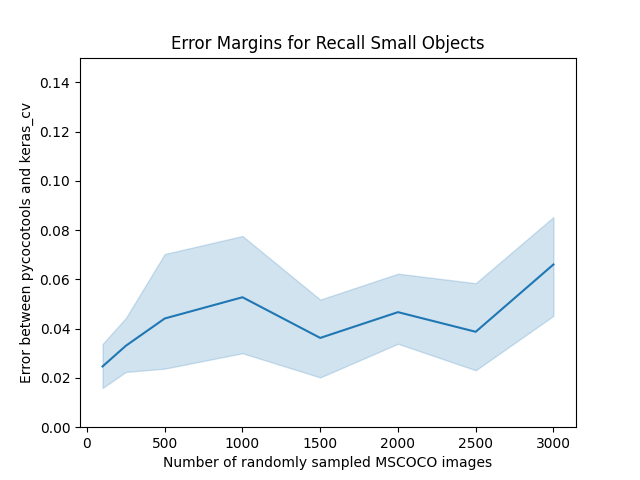}
        \end{minipage}
    \end{figure}

    \begin{figure}[H]
        \begin{minipage}{0.48\textwidth}
            \includegraphics[width=\textwidth]{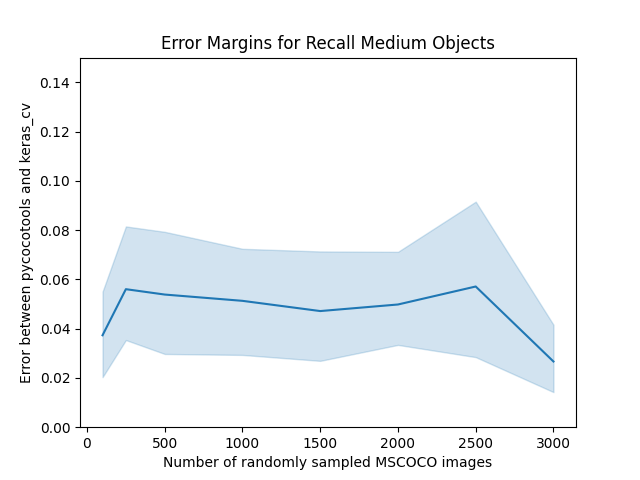}
        \end{minipage}
        \begin{minipage}{0.48\textwidth}
            \includegraphics[width=\textwidth]{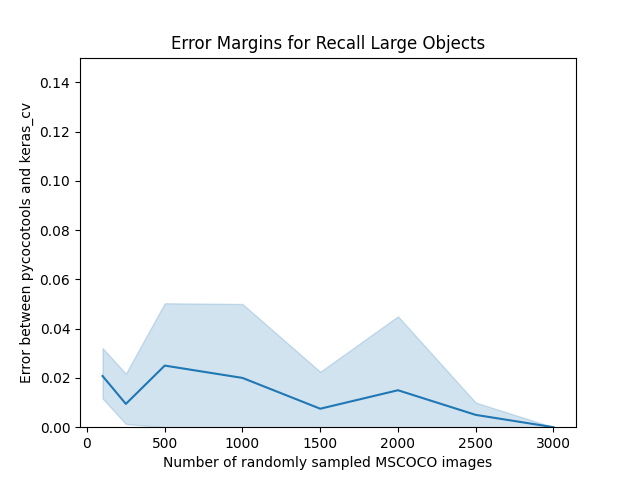}
        \end{minipage}
    \end{figure}
\end{minipage}
\caption{Error margins for each COCO recall challenge configuration}
\label{fig:error-margins-recall}
\end{figure*}

Figure \ref{fig:violin_plot} provides an overview of the error distributions of the runs of each metric.  Figure \ref{fig:error-margins-map} and Figure \ref{fig:error-margins-recall} show the resulting error margin plots generated from this experiment.
The X axis shows the number of images used in the 10 runs making up that datapoint, and the Y axis represents the
absolute error from the \texttt{pycocotools} metric.

These figures show that when both implementations are used to evaluated the same corpus
of images, the results are near identical.  However, there is clearly a difference taken
in the implementation of area range evaluation.

The error margins for both MaP and recall are <3\% for cases that do not involve area
range filtering. Table \ref{table:numerical_results} show the median, mean, and standard
deviation of each evaluation case.  In cases that do not involve area range filtering
the error margins are likely caused by rounding differences between the NumPy and
TensorFlow, rounding errors induced by differing order of operations between our
implementation and the \texttt{pycocotools} implementation, and the approximation algorithm used
to compute MaP.

\begin{table}
\centering
\caption{Numerical Results Across All runs}
\label{table:numerical_results}
\begin{tabular}{lrrl}
\toprule
               Metric &  Min Error &  Max Error &    Mean Error \\
\midrule
         Standard MaP &      0.000 &      0.046 & $0.015\pm0.011$ \\
          MaP IoU=0.5 &      0.000 &      0.075 & $0.024\pm0.019$ \\
         MaP IoU=0.75 &      0.001 &      0.079 & $0.025\pm0.017$ \\
    MaP Small Objects &      0.000 &      0.202 & $0.054\pm0.040$ \\
   MaP Medium Objects &      0.001 &      0.156 & $0.055\pm0.036$ \\
    MaP Large Objects &      0.000 &      0.150 & $0.028\pm0.030$ \\
   Recall 1 Detection &      0.000 &      0.029 & $0.009\pm0.007$ \\
 Recall 10 Detections &      0.000 &      0.035 & $0.011\pm0.007$ \\
      Standard Recall &      0.000 &      0.035 & $0.011\pm0.007$ \\
 Recall Small Objects &      0.001 &      0.143 & $0.043\pm0.031$ \\
Recall Medium Objects &      0.000 &      0.172 & $0.047\pm0.038$ \\
 Recall Large Objects &      0.000 &      0.150 & $0.013\pm0.032$ \\
\bottomrule
\end{tabular}
\end{table}

\subsection{Area Range Differences}

The error margin for all cases involving area filtering is significantly greater.  This
can be attributed to an implementation difference.  \texttt{pycocotools} filters the area of
bounding boxes according to the area annotation of the ground truth annotations included
in the COCO dataset.  It should be noted that in the COCO dataset the area of each
bounding box is actually computed using the segmentation map included in the dataset.
Additionally, no filtering is done on the predicted bounding boxes.

On the other hand, KerasCV filters bounding boxes according to the specified area range
by computing the area of the bounding box itself.  This is done for both the ground
truth bounding boxes and the detected bounding boxes.

While neither approach is perfect, we consider our approach to be significantly more idiomatic than that of \texttt{pycocotools}.  Segmentation maps are significantly more expensive to produce.  As such, most users performing bounding box detection will likely not have segmentation maps for the data they are evaluating their model on.  This would prevent these users from mirroring the approach taken in \texttt{pycocotools}.
Additionally, it is a confusing delegation of responsibilities to use a value computed using a segmentation map to evaluate a metric for bounding box detection.
Finally, tracking area of a segmentation map is simply a pain to the end user.  There is no general case reason for this behavior, and as such we have chosen to deviate from it.

\section{Conclusions}
We proposed and implemented an in-graph, mini-batch friendly approach to compute the COCO recall metric, as well as closely approximating the COCO mean average precision (MaP) metric.
We released an open-source, easy-to-use implementation of both algorithms as part of the KerasCV package.
Our numerical benchmarks confirm that our implementation produces results that closely match the most commonly-used prior implementation of both metrics, available in \texttt{pycocotools}.
With our in-graph approach, it becomes easy and efficient to implement train-time evaluation of object detection models.
This enables researchers and practitioners to leverage the full training curves for these metric to more effectively compare two models or prototype new models.

\printbibliography

\end{document}